\documentclass[conference]{IEEEtran}
\IEEEoverridecommandlockouts
\usepackage{cite}
\usepackage{amsmath,amssymb,amsfonts}
\usepackage{subfigure}
\usepackage{algorithm,algcompatible}
\usepackage{booktabs}
\usepackage{graphicx}
\usepackage{textcomp}
\usepackage{xcolor}
\usepackage{multirow}
\usepackage{svg}

\def\BibTeX{{\rm B/home/ubuntu/project/VITTRT/model.engine\kern-.05em{\sc i\kern-.025em b}\kern-.08em
    T\kern-.1667em\lower.7ex\hbox{E}\kern-.125emX}}

\begin{document}

\title{FAQS: Communication-efficient \underline{F}ederate DNN \underline{A}rchitecture and \underline{Q}uantization Co-\underline{S}earch for personalized Hardware-aware Preferences\\
{\footnotesize}
\thanks{}
}

\author{\IEEEauthorblockN{ Hongjiang Chen$^1$
, Yang Wang$^1$
, Leibo Liu$^1$
, Shaojun Wei$^1$
, Shouyi Yin$^1$}
\IEEEauthorblockA{\textit{
$^{1}$The School of Integrated Circuits, Tsinghua University, Beijing 100084, China.} \\
}
}

\maketitle

\begin{abstract}
Due to user privacy and regulatory restrictions, federate learning (FL) is proposed as a distributed learning framework for training deep neural networks (DNN) on decentralized data clients. Recent advancements in FL have applied Neural Architecture Search (NAS) to replace the predefined one-size-fit-all DNN model, which is not optimal for all tasks of various data distributions, with searchable DNN architectures. However, previous methods suffer from expensive communication cost rasied by frequent large model parameters transmission between the server and clients. Such difficulty is further amplified when combining NAS algorithms, which commonly require prohibitive computation and enormous model storage. Towards this end, we propose FAQS, an efficient personalized FL-NAS-Quantization framework to reduce the communication cost with three features: weight-sharing super kernels, bit-sharing quantization  and masked transmission. FAQS has an affordable search time and demands very limited size of transmitted messages at each round. By setting different personlized pareto function loss on local clients, FAQS can yield heterogeneous hardware-aware models for various user preferences.
Experimental results show that FAQS achieves average reduction of 1.58x in communication bandwith per round compared with normal FL framework and 4.51x compared with FL+NAS framwork.



\end{abstract}

\begin{IEEEkeywords}
Federate Learning, Neural Architecture Search, Quantization, Communication-efficiency
\end{IEEEkeywords}

\section{Introduction}
Federated Learning (FL) is a promising distributed training approach for deep neural networks (DNN) when  processing decentralized data due to user privacy and regulatory restrictions [1]. It allows multiple data client to collaboratively train  a global predifined model by exchanging intermediate parameters (e.g. the weights and biases) while keeping the raw local data samples private. As such, FL has been extensively applied in various challenging machine learning tasks such as data mining, computer vision and natural language processing.
Despite its widespread superiority, one major challenge involved in FL is data heterogeneity, which means that the data distributions across clients are not identically or independently (non-IID) in nature. 
As a result, it is difficult to manually design optimization schemes for pre-defined model architectures. 

\begin{figure}[htbp] 
\centering 
\includegraphics[width=0.49\textwidth]{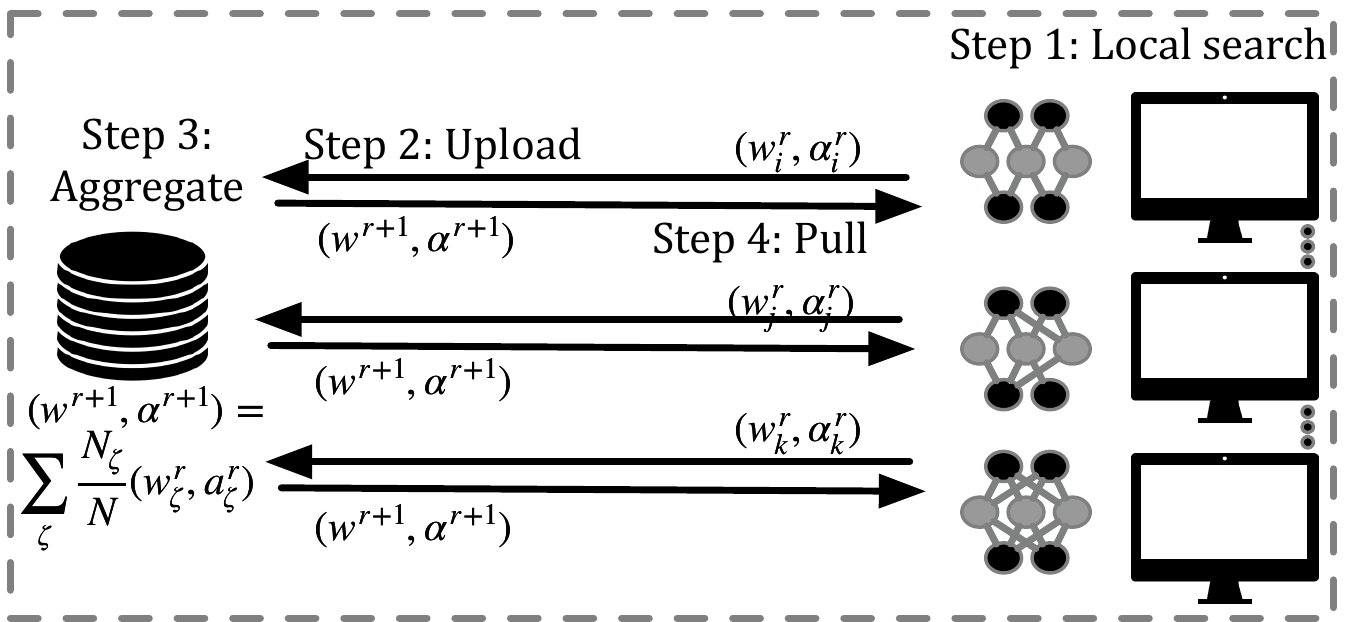} 
\caption{Framework of a normal FLNAS method on each round.} 
\end{figure}

In order to relax this constraint, recent research have combined Neural Architecture Search (NAS) algorithm to automatically generate high performance models instead of exploiting pre-defined architectures [2]-[5]. As showed in Fig. 1, the common NAS-based FL framework [2] can be divided into 4 steps. First, a NAS agent is ultizied as a client searcher to generate the network weights $w^{r}_{\zeta}$ and the neural architecture parameters $\alpha^{r}_{\zeta}$ with several epochs of local training at round $r$. Second, every client node uploads $w^{r}_{\zeta}$ and $\alpha^{r}_{\zeta}$ to the center server. 
Third, the server then aggregates all parameters to obtain a global version of $w^{r+1}$ and $\alpha^{r+1}$. At last, the clients pull the updated parameters for the following round of searching. 
However, there are two challenges in front of the current framework.  First, the NAS algorithms conducted on local clients require prohibitive computation and enormous model storage, which makes the parameters transmission extremely expensive because attempting many rounds of training on edge devices results in a remarkably huge communication cost between the server and clients. Second, it still searches for a unified global model, whih may not perform optimally on all local clients.

Based on the previous observations, we propose \textbf{FAQS}, an efficient personalized \textbf{F}L-N\textbf{A}S-\textbf{Q}uantization Co-\textbf{S}earch framework to reduce the communication cost in three dimensions. \textbf{First}, we adopt the single-path based NAS algorithm [12] as our searching agent featuring weight-sharing super kernel, which slims the redundant supernet to the similar size of a normal network. \textbf{Second}, the network parameters is further compressed  with mixed-precision bit-sharing quantization, which is jointly conducted with neural architecture search to save extra search time expense. \textbf{Third}, only a part of the local network is transmitted using masked parameters when uploading or pulling to save considerable size of communications. All these factors contribute to a communication-efficient manner of generating personalized local models.
In addition, FAQS can satisfy heterogeneous hardware-aware preferences by setting various personlized pareto functions on different local clients.

Our contributions of this work can be concluded as follows:
\begin{itemize}
\item We propose a communication-efficient personalized FL-NAS-Quantization framework to reduce the transmission cost between the server and clients. This is the first work (to the best of our knowledge) to joint optimize NAS and Quantization policy in a FL framework.
\item We diagnose and lift three techniques featuring \textbf{weight-sharing super kernel}, \textbf{bit-sharing quantization} and \textbf{masked transmission} to achieve average reduction of 1.58x in communication bandwith per round compared with normal FL framework and 4.51x compared with FL+NAS framwork.
\item We demonstrate that FAQS can yield heterogeneous hardware-aware models for various user preferences by setting different personlized pareto functions on accuracy, latency and model size.
\end{itemize}
\section{Motivation}
\subsection{Differential Neural Architecture Search for FL}
Great research interests have been raised in Neural Architecture Search for automatically genereting high-performance DNNs without human bias. Current NAS algorithms is typically based on three methods: Reinforcement Learnning (RL) [6]-[10], Evolutionary Algorithm (EA) [16], [17] and Gradient Differential supernet (GD) [11]-[15]. Both RL and EA require an enormous searching time to find the best operations due to ineffective search direction proposals. While GD-based methods dramatically decreases the search time by formulating a differentiable search space, which has obtained momentum in recent research of FL to search for efficient DNNs. Direct Federated NAS [3] ultizes DSNAS [21] with Fed averaging algorithm in search of a unified model.  FedNAS [2] combines MileNAS [13] with Fed averaging algorithm to search for a global model. [22] explores the concept of differential privacy with DARTs [11] to analyze the trade-off between accuracy and privacy of a global model. However, all these models are faced with two challenges. First, prior research is based on a differentiable multi-path  NAS algorithm which searchs over a supernet that encompasses all candidate architecture paths. Although it reduces the total search epochs to acceptable quantity, the number of candidate paths grows exponentially w.r.t. the number of hyper-parameter types which makes the whole supernet overweighted. As a result, every client needs to transmit tremendous size of the local model to the center server on each round and thus leads to huge communication cost. Second, existing methods converge on a single unified model's architecture and weight parameters. As a result, there can only be one predictive outcome which may not perform optimally on all local clients' tasks.


\subsection{Joint optimization on quantization and NAS }
In order to achieve higher energy efficiency with limited resource budget, DNNs must be carefully designed in two steps: the architecture design and quantization policy choice. However, taking the two steps separately is time-consuming and leads to a sub-optimal final deployment [20]. Recent research have attempted joint optimization for both quantizaiton and NAS. [20] ultizes a RL agent based on a multi-objective evolutionary search algorithm to search the models under the balance between model size and performance accuracy. [19] formulates the co-search problem by fusing DNN search variables and hardware implementation (quantization and other implementation) variables into one solution space, and maximize both algorithm accuracy and hardware implementation quality. [18] proposes a novel co-exploration framework which parameterizes the layer-wise quantization and searches these parameters jointly with the hyperparameters of the architecture. However, prior methods are lacking in searching efficiency. [18] and [20] simply wrap an outer loop of the original NAS framework to handle the quantization selection and [19] encompasses quantization policy as another multi-path dimension which increases both the time of search and the model size.
\section{FAQS Framework}
In this section, we introduce our entire framework FAQS, an efficient personalized FL-NAS-Quantization Co-Search framework to reduce the communication cost between the server and clients. In the rest part of this section, we first introduce the problem definition including FAQS problem formulation and the communication cost. Next we lift and explain three key techniques used in our framework to reduce communication cost in different dimensions. Last, we give an overall explaination on FAQS algorithm. 
\begin{figure*}[htbp] 
\centering 
\includegraphics[width=1\textwidth]{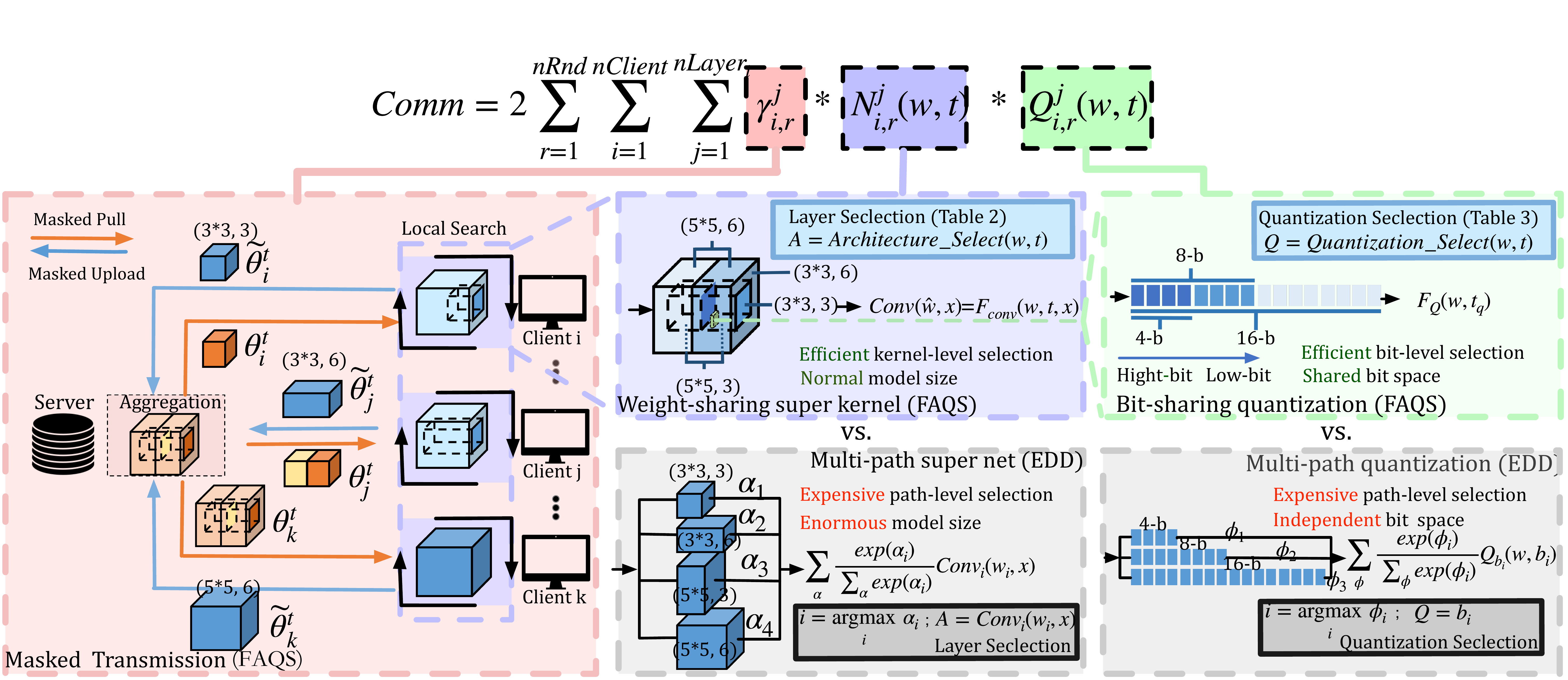} 
\caption{FAQS framework: Masked transmission (left), Weight-sharing super kernel (top middle) and Bit-sharing quantization (top right). Comparison with EDD: Multi-path super net (bottom middle) and Multi-path quantization (bottom right).} 
\end{figure*}
\subsection{Problem Definition}
\textbf{FAQS problem formulation}: Mathematically, FAQS can be represented by the following optimization problem. In FAQS, the objective is to find a personalized model for each device $i$ that performs well on the local data distribution (we assume each client has the same number of samples for simplicity):
\begin{equation}
\setlength{\abovedisplayskip}{0pt}\setlength{\belowdisplayskip}{0pt}
\mathop{\min}_{\theta_{1},..,\theta_{k}} G(\theta_{1},..,\theta_{k})=\frac{1}{k}\sum_{i=1}^{k}L_{i}(\theta_{i},x_{i},y_{i})
\end{equation}
\begin{equation*}
\begin{split}
\setlength{\abovedisplayskip}{0pt}\setlength{\belowdisplayskip}{0pt}L_{i}(\theta_{i},x_{i},y_{i}) = \alpha_{i}CE(\theta_{i},x_{i},y_{i})+\beta_{i}&Lat(\theta_{i})+\gamma_{i}MS(\theta_{i})\\
\alpha_{i}+\beta_{i}+\gamma_{i}&=1
\end{split}
\end{equation*}
Where $k$ is the number of clients and $\theta_{i}$ is the trainable parameters including weights $w$ and architecture thresholds $t$. $(x_{i},y_{i})$ represents local data of each client. $L_{i}$ is the personalized loss function of $client_{i}$ with soften pareto coefficient [19] $(\alpha_{i},\beta_{i},\gamma_{i})$ on accuracy (converged by cross-entropy loss), latency and model size. We utilize the differentiable $Lat$ and $MS$ loss used in [27], which regards the weighted sum of latency or model size on each type of block, as final $Lat$ and $MS$ result for end-to-end training.

\textbf{Communication Cost}: The total communication cost (bits) transmitted between the server and clients through all rounds in FAQS can be formulated as:
\begin{equation}
\setlength{\abovedisplayskip}{0pt}\setlength{\belowdisplayskip}{0pt}
Comm=2\sum_{r=1}^{nRnd}\sum_{i=1}^{nClient}\sum_{j=1}^{nLayer_{i}}\gamma_{i,r}^{j}N_{i,r}^{j}(w,t)*Q_{i,r}^{j}(w,t)
\end{equation}
Where $nRnd$ represents the total rounds of FL, $nClient$ is the number of clients involved in the collaborative training and $nLayer_{i}$ is the $i_{th}$ client's total layer number. $w$ and $t$ are learnable weights and architecture parameters respectively. FAQS generates specific architectures based on thresholds $t$ instead of the weighted path-level coefficient $\alpha$ used in muti-path [2]. 
$\gamma_{i,r}$ represents the proportion of the entire network parameters which are not masked during message transmimission. $N_{i,r}^{j}(w,t)$ is the total number of parameters including weights (or gradients) and architecture parameters while $Q_{i,r}^{j}(w,t)$ represents the number of bits on $layer_{j}$ of $client_{i}$ in $round_{r}$. Since each round catains parameters pulling (from the server to clients) and uploading (from clients to the server), the commuication cost is times by 2 for the final result. For a better understanding on previous parameters, we compare FAQS with FedAvg [26]  and FedNAS+EDD [2], [19] (our implement of EDD with FedNAS) on each variable. As showed in Table \uppercase\expandafter{\romannumeral1}, we focus on a single muti-path MobileNetV2 block [23] which is the basic searchable unit in [2] and [19]. By adapting single-path weight-sharing super kernel and bit-sharing quantization, both the number of total parameters and the average 
quantization bits are reduced in FAQS. $\gamma_{i,r}^{j}$ represents the percentage of the parameters that are actually transmitted. FedAvg and FedNAS-EDD both transmit the entire network for a full traning, while FAQS 
masks part of the network inspired by [25]. In other words, FAQS only transmit the crucial parts that affect the gradient aggregation hence $\gamma_{i,r}^{j}$ can be decreased.

\begin{table}[htbp]
\caption{Parameters comparison on different methods(evaluated with a single MBV2-like searching block with 24 input channels and expand ratio=$\{3,6\}$ on the same client)}
\normalsize
\centering
\scalebox{0.69}{
\begin{tabular}{cccccc}
\toprule  
Method&$\gamma_{i,r}^{j}$&$N_{i,r}^{j}(w,t)$&$Q_{i,r}^{j}(w,t)$&$Comm_{i,r}^{j}$(KB)&Personalized \\
\midrule 
FedAvg(MBV2)&1&9k&32(float)& 72&no\\
FedNAS+EDD&1&25k&28(quant)&175&yes\\
FAQS(ours)&0.51&9k&16(quant)&18.4&yes\\
\bottomrule 
\end{tabular}
}
\end{table}

\subsection{Technique Details On Communication Reduce}

As showed in Fig. 2, the framework of FAQS adpats three techniques to reduce diverse parameters in Eq. 2 for a communication-efficient manner of generating personalized local models including weight-sharing super kernel (top middle), bit-sharing quantization (top right) and masked transmission (left).

\textbf{Weight-sharing super kernel}: In this dimension, we reduce the total number of parameters $N_{i,r}(w,t)$ in Eq. 2 Compared with the famous differential multi-path NAS [11] (showed in Fig. 2 bottom middle), which constructs and directly trains an over-parameterized network named SuperNet containing all possible candidate paths. The key idea of single-path NAS is relaxing architecture decisions over a $\textbf{weight-sharing over-parameterized kernel}$ called super kernel [12]. A super kernel is the max size of the potential kernel. For MobilenetV2, it is a $5*5$ kernel with the expansion ratio of 6. As showed in Fig. 2 (top middle), we divide the whole kernel in several parts. For searching kernel size, choosing $3*3$ or $5*5$ is equivalent to whether we choose the subset of $w_{55\backslash33}$ or not. And for searching expand ratio and skip opertaion, it is determined by if we choose to preserve the entire length, half length or zero of the kernel. 
\begin{equation}
\setlength{\abovedisplayskip}{1pt}\setlength{\belowdisplayskip}{1pt}
w_k=w_{33}+\sigma(id(w_{55\backslash33},t_{k,5}))*w_{55\backslash33}
\end{equation}
\begin{equation}
\setlength{\abovedisplayskip}{1pt}\setlength{\belowdisplayskip}{1pt}
\hat{w}=\sigma(id(w_{k,3},t_{e,3}))*(w_{k,6\backslash3}+\sigma(id(w_{k,3},t_{e,6}))*w_{k,6\backslash3})
\end{equation}

Where $w_{33}$ and $w_{55\backslash33}$ represet weights of the 3*3 part and 5*5$\backslash$3*3 part of the orginal super kernel. $w_k$ is the kernel shape selection output. While $w_{k,3}$ and $w_{k,6\backslash3}$ represent the first and the second half of the expanded super kernel. $t_{k,5}$, $t_{e,3}$ and $t_{e,6}$ are three thresholds fed on three $\sigma(\cdot)$ sigmoid functions to decide whether we choose the subset of $w_{55\backslash33}$, the skip operation and the expand ratio of 3. The original problem is a hard-decision problem which can be softened by using these sigmoid functions. The $id(\cdot)$ function is used to caculate the indicator value as input of the sigmoid. We simply use the group Lasso term of the corresponding weight values and a threshold to make our decisions. For example, towards the following equation:
\begin{equation*}
\setlength{\abovedisplayskip}{3pt}\setlength{\belowdisplayskip}{3pt}
id(w_{55\backslash33},t_{k,5})=\parallel w_{55\backslash33}\parallel^2-t_{k,5}
\end{equation*}
We can duduce from the the $id(\cdot)$ result that if $id(\cdot)$ is much bigger than zero, then most variables in the $w_{55\backslash53}$ are likely to have important values hence $\sigma{(\cdot)}$ is almost equal to 1. As a result $w_k$ yield a complete $w_{55}$ kernel size according to Eq. 2. By using this single-path NAS based on weight-sharing super kernels, the original redunt supernet in multi-path NAS can be slimmed to the similar size of a normal network, making the entire searh agent light and efficient. The relationship between the block type and $id(\cdot)$ result is enumerated in Table \uppercase\expandafter{\romannumeral2}.
\begin{equation*}
\setlength{\abovedisplayskip}{1pt}\setlength{\belowdisplayskip}{1pt}
output=Conv(\hat{w},x)=F_{conv}(w,t,x)
\end{equation*}
The final output of this layer can be caculated as a convolution on transformed weight $\hat{w}$ and input $x$ or as a general function $F_{conv}$ on original weight $w$, architecture threshold $t$ and $x$.
\begin{table}[htbp]
\caption{Mapping from $id(\cdot)$ functions to architecture block selection ($+/-$ represent positive/negtive values and $*$ represents any value)}
\normalsize
\centering
\scalebox{0.8}{
\begin{tabular}{cccc}
\toprule  
$id(w_{k,3},t_{e,3})$&$id(w_{k,3},t_{e,6})$&$id(w_{55\backslash33},t_{k,5})$ &$block(size, ratio)$\\
\midrule 
$+$&$+$&$+$& $5*5, 6$\\
$+$&$+$&$-$& $3*3, 6$\\
$+$&$-$&$+$& $5*5, 3$\\
$+$&$-$&$-$& $3*3, 3$\\
$-$&$*$&$*$& $skip$\\
\bottomrule 
\end{tabular}
}
\end{table}

\textbf{Bit-sharing quantization}:
In this dimension, we focus on decreasing the average number of bits on all quantized variables $Q_{i,r}^{j}(w,t)$ in Eq. 2 when transmitting messages between the server and the clients. Quantization is a model compression method when deploying DNNs. We follow the widely used quantization aware training method [24], which uniformly round floating point values to the nearest integer point values based on the maximum and the minimum representative values. We fomulate the quantization process as follows:
\begin{equation}
\setlength{\abovedisplayskip}{1pt}\setlength{\belowdisplayskip}{1pt}
w^*=Norm^{-1}(R(Norm(w), b))=Q(w,b)
\end{equation}
Where $(w, w^*)$ is the weight vector of the real value and the approximate quantization value, and $b$ is the bit number. $Norm(w)=\frac{w-w_{min}}{w_{max}-w_{min}}$ is a linear scaling functiong which normalizes the values of the vectors into [0,1] and $Norm^{-1}$ is the inversed function. In paticular, $Q(\cdot)$ is the quantization function which takes in the normalized real value $w$ and the bit number $b$ and outputs the closest quantization value of the corresponding weight element. Different from the differencial multi-path quantization policy in [19] that allocates independent bits for diverse quantization policy storage (Fig. 2 bottom right), FAQS stores \textbf{all quantization policies in shared space}. For example, for a $\{4,8,16\}$ bit search space, [19] takes 28$(4+8+16)$ bits to store each potential quantization choice independently while FAQS only needs 16 fixed bits (b=16). As a result, 42\% of model size can be saved on communications in FL. 
As showed in Fig. 2 (top right), instead of selecting quantization policy out of independent multi-path quantization policy, the basic idea of bit-sharing is similar to weight-sharing super kernels. For each layer, the quantization policy is decided by whether we choose to preserve the 5$\sim$8 bits and the 9$\sim$16 bits in a fixed 16-bit space or not.
\begin{equation}
\begin{split}
\setlength{\abovedisplayskip}{1pt}\setlength{\belowdisplayskip}{1pt}
w_q=&w_{1\sim4}^*+\sigma (id(w_{5\sim8}^*,t_{q,5\sim8})) \\
&*(w_{5\sim8}^*+\sigma (id(w_{9\sim16}^*,t_{q,9\sim16}))*w_{9\sim16}^*) \\=&f(w^*,t_{q})=f(Q(w,16),t_{q})=F_{Q}(w,t_{q})
\end{split}
\end{equation}
Where $w_{i\sim j}^*$ is the $i_{th}$ to $j_{th}$ bits of the approximate quantization value in Eq. 5 (left to right: high bit to low bit). Similarly, $t_{q,5\sim8}$ and $t_{q,9\sim16}$ are two thresholds as inputs for respective $\sigma(\cdot)$ functions to determine whether we preserve the 5$\sim$8 bits and the 9$\sim$16 bits. Similar to the former weight-sharing super kernels, towards the following $id(\cdot)$ function:
\begin{equation*}
\setlength{\abovedisplayskip}{1pt}\setlength{\belowdisplayskip}{1pt}
id(w_{5\sim8}^*,t_{q,5\sim8})=\parallel w_{5\sim8}^*\parallel^2-t_{q,5\sim8}
\end{equation*}
If $\parallel w_{5\sim8}^*\parallel^2$ is much bigger than the threshold $t_{q,5\sim8}$, most weights' $5\sim8$ bits are non-zeros. As a result, an 8-bit(at least) quantization policy is needed for the fine-grained weight distribution. In contrast, when most $5\sim8$ bits are zeros, we can approxiamtely regard the original 16-bit weight as a 4-bit ($1\sim4$ bits) manner. The relationship between quantization policy and $id(\cdot)$ result is enumerated in Table \uppercase\expandafter{\romannumeral3}.
\begin{table}[htbp]
\caption{Mapping from $id(\cdot)$ functions to quantization bits choice ($+/-$ represent positive/negtive values and $*$ represents any value)}
\normalsize
\centering

\scalebox{0.9}{
\begin{tabular}{ccc}
\toprule  
$id(w_{5\sim8}^*,t_{q,5\sim8})$&$id(w_{9\sim16}^*,t_{q,9\sim16})$&$Quantization(bit)$\\
\midrule 
$+$&$+$&$16$\\
$+$&$-$&$8$ \\
$-$&$*$&$4$ \\
\bottomrule 
\end{tabular}
}
\end{table}



\begin{algorithm}
\caption{FAQS} 
\label{alg1}
\begin{algorithmic}
\renewcommand{\algorithmicrequire}{\textbf{Input:}}
\renewcommand{\algorithmicensure}{\textbf{Output:}}
\REQUIRE local searching epochs $E$; learning rounds $T$; clients number $K$; local training set $D^{train}_{i}$ and personalized pareto coefficients $(\alpha_{i},\beta_{i},\gamma_{i})$ for each client ${i}$.

\ENSURE Local model $P_{i}$ with architecture $A_{i}$, quantization $Q_{i}$ and Weight $W_{i}$
\STATE $Main\_Function:$
\STATE $\theta^{0}_{i} \leftarrow Initialize\_Parameter$  
\STATE \textbf{For} {$t = 1,2..T$} \textbf{Do}
\STATE {\ \ \ \ }\textbf{For} {$i = 1,2..K$ \textbf{in parallel Do}}
\STATE {\ \ \ \ \ \ \ \ } $\widetilde\theta^{t}_{i} \leftarrow Local\_Search(\theta^{t-1}_{i},D^{train}_{i},\alpha_{i},\beta_{i},\gamma_{i})$
\STATE {\ \ \ \ }$(\theta^{t}_{1},..,\theta^{t}_{k})\leftarrow Aggregation\&Masked\_Pull(\widetilde\theta^{t}_{1},..,\widetilde\theta^{t}_{k})$
\STATE \textbf{For} {$i = 1,2..K$ \textbf{in parallel Do}}
\STATE {\ \ \ \ }$P_{i}\leftarrow Local\_Finetune(\theta^{T}_{i},D^{train}_{i})$

\STATE
\STATE $Local\_Search(\theta, D^{train},\alpha,\beta,\gamma):$
\STATE $w,t\leftarrow \theta$
\STATE \textbf{For} {$e = 1,2..E$} \textbf{Do}
\STATE {\ \ \ \ }\textbf{For} $batch(x,y)\in D^{train}$ \textbf{Do}
\STATE {\ \ \ \ \ \ \ \ }$w_{q} \leftarrow Bit\_Sharing\_Quantizate(w,t)$ (Eq. 6,7)
\STATE {\ \ \ \ \ \ \ \ }$\hat{w} \leftarrow Weight\_Sharing\_Kernel(w_{q},t)$ (Eq. 4,5)
\STATE {\ \ \ \ \ \ \ \ }$w \leftarrow w - \eta_{w}\nabla_wL(batch(x,y),\alpha,\beta,\gamma, \hat{w})$
\STATE {\ \ \ \ \ \ \ \ }$t \leftarrow t - \eta_{t}\nabla_tL(batch(x,y),\alpha,\beta,\gamma,\hat{w})$
\STATE $\widetilde{w} \leftarrow Masked\_Sample(w)$ (Table \uppercase\expandafter{\romannumeral2})
\STATE $\widetilde{\theta}\leftarrow \widetilde{w},t$
\STATE \textbf{Masked Upload} $\widetilde{\theta}$ \textbf{to server}

\STATE
\STATE $Local\_Finetune(\theta, D^{train}):$
\STATE $w,t\leftarrow \theta$
\STATE $\widetilde{W} \leftarrow Masked\_Sample(w)$ (Table \uppercase\expandafter{\romannumeral2})
\STATE $A\leftarrow Architecture\_Select(w,t)$(Table \uppercase\expandafter{\romannumeral2})
\STATE $Q\leftarrow Quantization\_Select(w,t)$(Table \uppercase\expandafter{\romannumeral3})
\STATE $W\leftarrow Train(A,Q,\widetilde{W},D^{train})$
\STATE $P\leftarrow (A,Q,W)$
\STATE \textbf{Return} $P$

\end{algorithmic}
\end{algorithm}

\textbf{Masked transmission}: In this part, we seek the oppotunity to reduce the valid proportion $\gamma_{i,r}^{j}$ of network parameters transmission in Eq. 2. Inspired from FedMask [25] which masks unimportant channels and only aggregates the overlapped weights, the basic idea behind masked transmission is similar: we only \textbf{transmit the important parts of kernels} generated by local search agents. For example, as showed in Fig. 2 (left): towards a paticular layer in round $r$, masked transmission means that we do not pull or upload the entire MBV2 block with regressed values caculated by Eq. 3 and Eq. 4. Instead, discrete architectures are selected according the mapping relationship in Table \uppercase\expandafter{\romannumeral2}. In this case, $client_{i}$ $client_{j}$ and $client_{k}$ select $(3*3, 3)$, $(3*3, 6)$ and $(5*5, 6)$ MBV2 blocks respectively. Where $(k*k,e)$ represents the MBV2 block shape with kernel size $k$ and expand ratio $e$. Then FAQS only transmits the part of blocks with respective kernel shape and masks the other unimportant values. In this way, large amount of kernel weights can be left out and the valid proportion $\gamma_{i,r}^{j}$ of network parameters can be reduced to small numbers. When it comes to weights aggregation, only corresponding overlapped shape of kernels are calculated for aggregation. For example, the $(3*3, 3)$ part is shared among all these clients while the $(3*3, 6\backslash3)$ part accounts for $client_j$ and $client_k$. And the $(5*5\backslash3*3, 6)$ part only accounts for $client_k$. As a result, $(3*3, 3)$ is aggregated for all clients and the $(3*3, 6\backslash3)$ part is pulled for $client_j$ and $client_k$. while $(5*5\backslash3*3, 6)$ is monopolized by $client_k$. Since it is not required to  transmit the complete block for all clients on each round, $\gamma_{i,r}^{j}$ is greatly is reduced for a communication-efficient FL framework.

\subsection{Overall Algorithm}

As showed in Algorithm 1, FAQS is generally a two-step algorithm. First, a two-depth nested loop is adapted to generate personalized local models with a communication-efficient manner. Second, after the final epoch of local search and server aggregation, each local model is discretely selected for finetuning to yield the final model $P_{i}$ with architecture $A_{i}$, quantization $Q_{i}$ and Weight $W_{i}$. To be more specific, FAQS fisrtly initialize each client $i$ with trainable parameters $\theta_{i}^{0}$ including weights $w_{i}^{0}$ and architecture thresholds $t_{i}^{0}$. For each round, every client conducts local joint search on neural architecture and quantization. The process of $Local\_search$ function is simple. The original kernel weights is firstly quantized in a fixed 16-bit format using Eq. 5 and Eq. 6. Then the output quantized weights is transformed into a regressed combination of different parts of kernels using Eq. 3 and Eq. 4 (denoted as $\hat{w}$). To this end, we regard $\hat{w}$ as the direct form of kernel weights where $output=Conv(\hat{w},x)$. The personalized batch loss is a function parameterized by $\hat{w}$ and pareto coefficients $(\alpha_{i},\beta_{i},\gamma_{i})$, where $\hat{w}$ is parameterized by $w$ and $t$. As a result, we can directly conduct mini-batch gradient descent on $w$ and $t$. Afterwards, we discretely select the masked weight $\widetilde{w}$ of each layer according to Table \uppercase\expandafter{\romannumeral2} and transmitt them to the server for aggregation. When finishing the final round of local search, FAQS generates the architecture and quantization selection based on the current $w$ and $t$ according to Table \uppercase\expandafter{\romannumeral2} and Table \uppercase\expandafter{\romannumeral3}. The finetune checkpoint starts on the masked sample weight $\widetilde{W}$ and the final local model $P_{i}$ is generated with architecture $A_{i}$, quantization $Q_{i}$ and Weight $W_{i}$.

\begin{figure*}[htbp]
\centering
\begin{minipage}[t]{0.3614\linewidth}
\centering
\includegraphics[width=1.\textwidth]{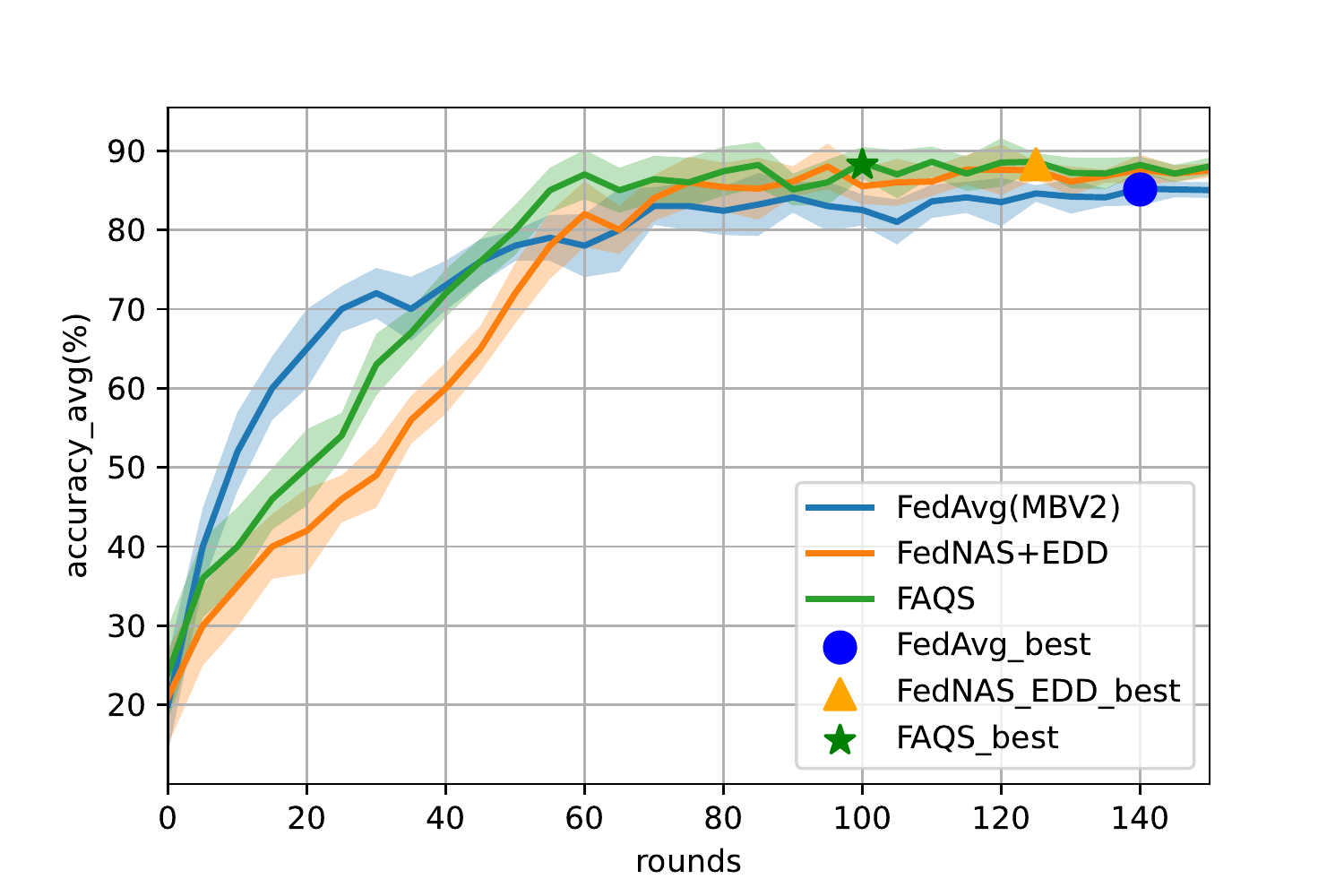}
\end{minipage}%
\begin{minipage}[t]{0.3614\linewidth}
\centering
\includegraphics[width=1.\textwidth]{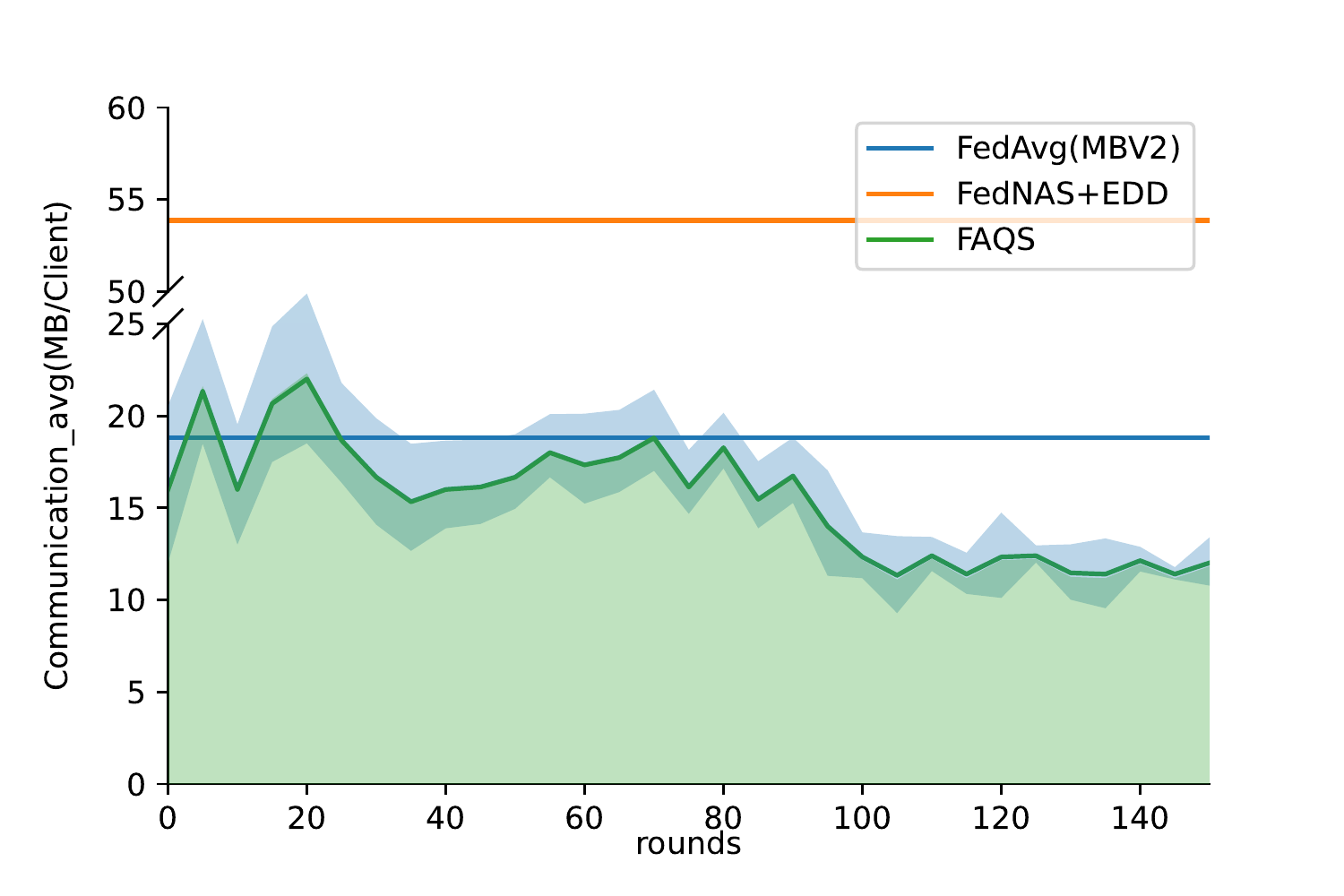}
\end{minipage}%
\begin{minipage}[t]{0.2771\linewidth}
\centering
\includegraphics[width=1.\textwidth]{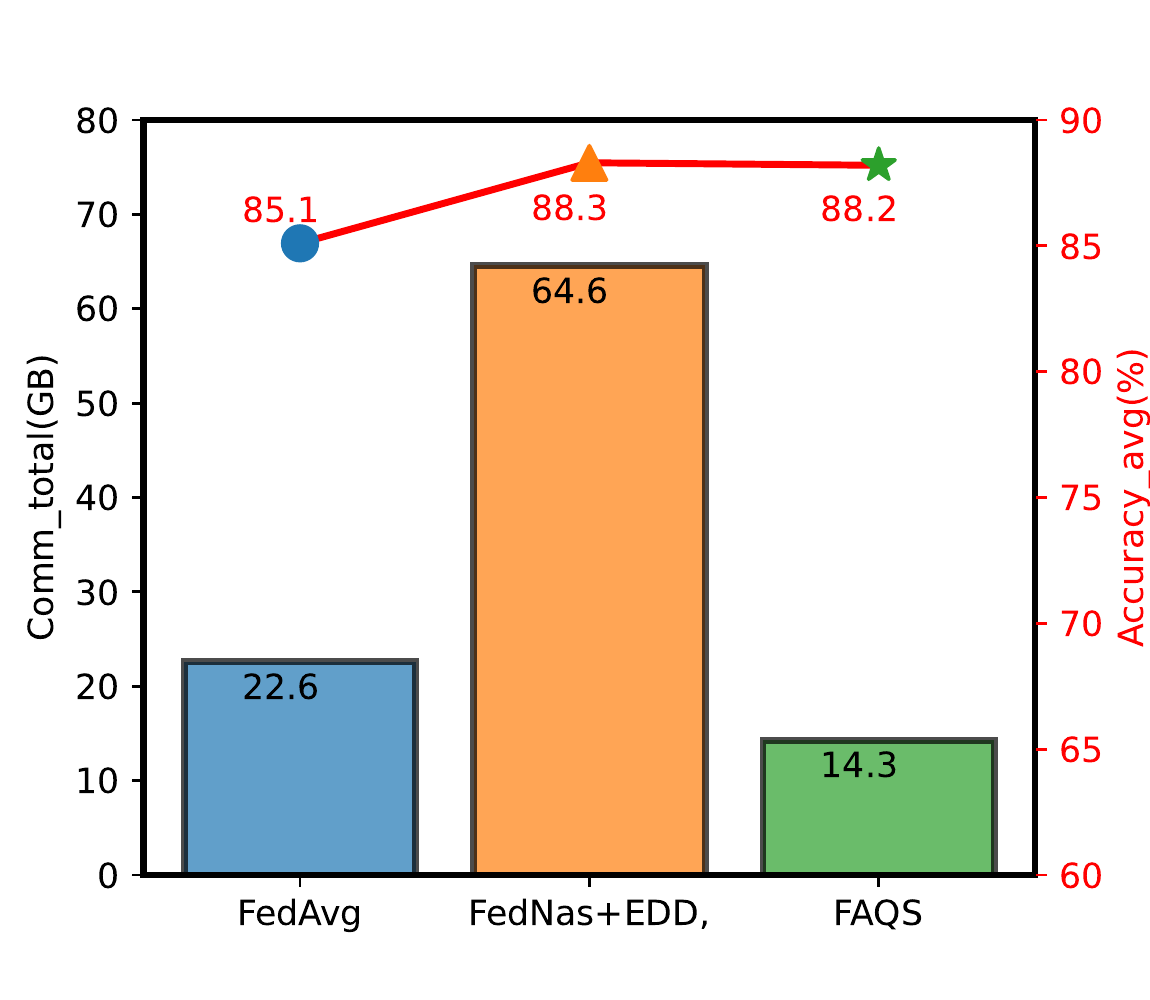}
\end{minipage}
\centering
\caption{Average accuracy (left) and average communication cost (middle) on each round; Communication and accuracy comparison (right) on FedAvg, FedNAS+EDD and FAQS.}
\end{figure*}

\begin{table*}[htbp]
\caption{Personalized Models for different Hardware-aware Preferences (R represents the downsampled layer)}
\normalsize
\centering

\scalebox{0.625}{
\begin{tabular}{ccccccccccccccccccccc}
\toprule  
\multicolumn{2}{c}{Model}& $L_{1}(R)$ & $L_{2}$ & $L_{3}$ & $L_{4}$ & $L_{5}(R)$ & $L_{6}$ &$L_{7}$&$L_{8}$&$L_{9}(R)$&$L_{10}$&$L_{11}$&$L_{12}$&$L_{13}(R)$&$L_{14}$&$L_{15}$&$L_{16}$ &$Acc(\%)$& $Lat(ms)$ & $MS(MB)$\\
\midrule 
\multicolumn{2}{c}{$FedAvg_{float}(MBV2)$} & (3,6) & (3,6) & (3,6) & (3,6) & (3,6) & (3,6) &(3,6)&(3,6)&(3,6)&(3,6)&(3,6)&(3,6)&(3,6)&(3,6)&(3,6)&(3,6) &85.1& 8.9 & 19.2\\

\midrule 
\multirow{2}{*}{$FAQS_A(0.8,0.1,0.1)$} &$(k,e)$&(3,6)&(3,3)&(3,3)&(3,3)&(5,6)&(5,3)&(3,3)&(3,3)&(5,6)&(5,3)&(3,3)&(3,3)&(5,6)&(5,6)&(3,6)&(3,6) &\multirow{2}{*}{\textbf{88.2}}&\multirow{2}{*}{8.6}& \multirow{2}{*}{15.1}\\
&$q$&16&16&16&8&16&16&8&8&16&16&16&8&16&8&16&16 & ~ & ~ & ~\\

\midrule 
\multirow{2}{*}{$FAQS_B(0.4,0.5,0.1)$} &$(k,e)$&(5,6)&(5,3)&skip&skip&(5,6)&(5,6)&skip&skip&(5,6)&(5,6)&skip&skip&(5,6)&(5,6)&(5,6)&(5,6) &\multirow{2}{*}{83.6}&\multirow{2}{*}{\textbf{5.2}}& \multirow{2}{*}{15.4}\\
&$q$&16&16&$\backslash$& $\backslash$&16&16&$\backslash$&$\backslash$&16&16&$\backslash$&$\backslash$&16&16&16&16 & ~ & ~ & ~\\

\midrule 
\multirow{2}{*}{$FAQS_C(0.4,0.1,0.5)$} &$(k,e)$&(3,6)&(3,3)&(3,3)&skip&(3,6)&(3,6)&skip&(3,3)&(3,6)&(3,3)&(3,3)&(3,3)&(3,6)&(3,6)&(3,6)&(3,6) &\multirow{2}{*}{86.1}&\multirow{2}{*}{7.8}& \multirow{2}{*}{\textbf{11.3}}\\
&$q$&16&16&16& $\backslash$&16&8&$\backslash$&8&16&8&8&8&16&8&16&16 & ~ & ~ & ~\\
\bottomrule

\end{tabular}
}

\end{table*}

\section{Evaluations}
Experiments are conducted on the standard classification benchmark to evaluate our method. We implement FAQS for distributed computing with nine nodes each equipped with a GPU (NVIDIA Tesla V100). We set up our experiment in a cross-silo setting for simplicity with one node representing the server and eight nodes representing the clients.

\subsection{Experimental Setup}
\textbf{Dataset preparation}: We select CIFAR10 dataset on image classification task for a fair comparison, which is commonly used as a benchmark in FL and NAS [2], [19]. We generate non-IID data across clients by exploiting LDA distribution [2] with parameter ($\alpha$=0.2). 

\begin{figure}[htbp]
\centering
\begin{minipage}[t]{0.5\linewidth}
\centering
\includegraphics[width=1.1\textwidth]{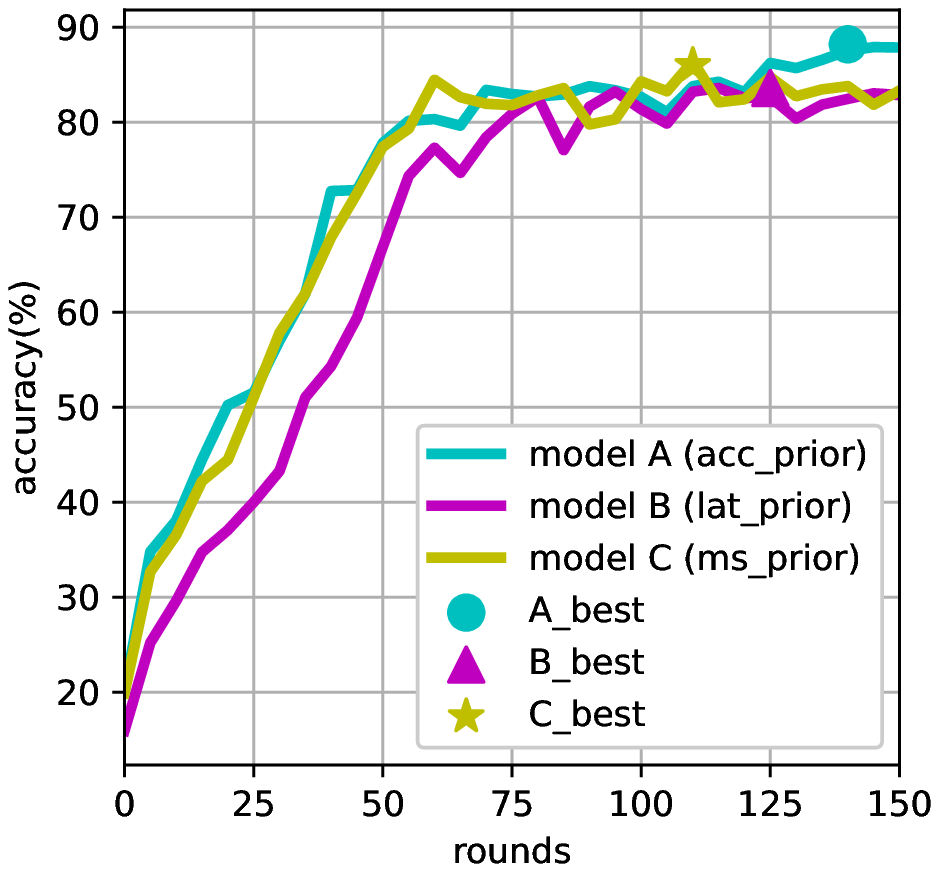}
\end{minipage}%
\begin{minipage}[t]{0.5\linewidth}
\centering
\includegraphics[width=1.1\textwidth]{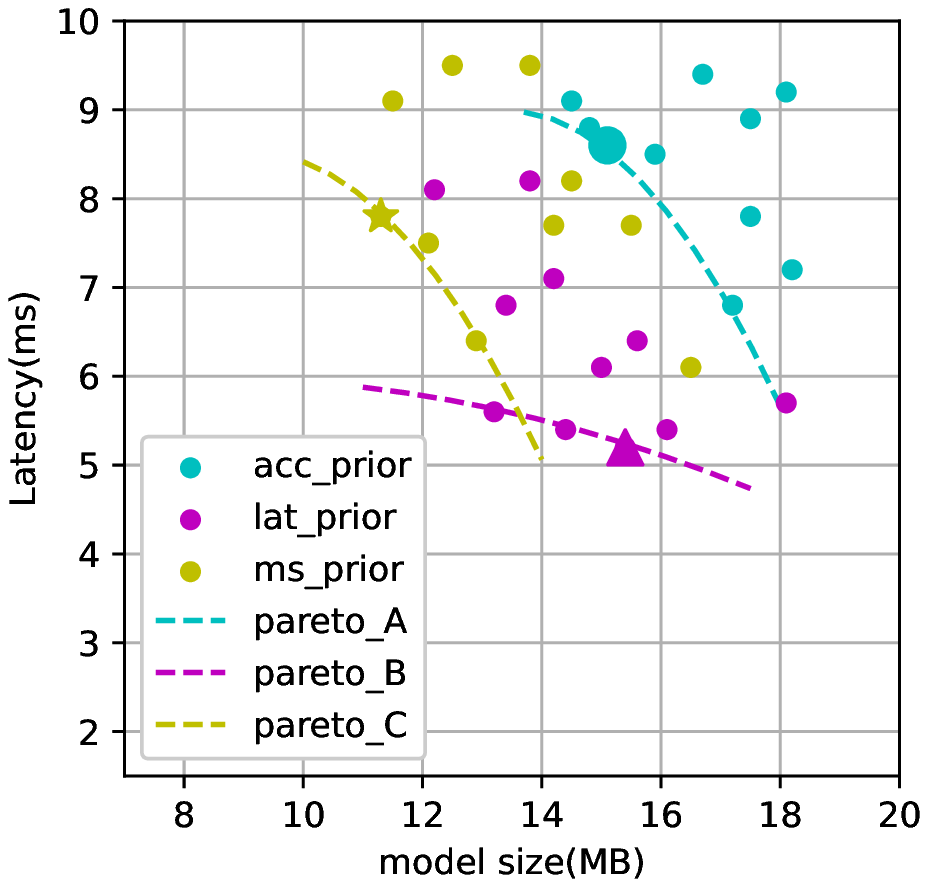}
\end{minipage}%
\centering
\caption{Traning accuracy curve (left) and Lat\_MS scattering (right) for  clients with Personal Hardware-aware Preferences.}
\end{figure}

\textbf{Search space}: For neural architecture search space, our framework builds upon hierarchical MBV2-like search space used in [23], [12], [14], [19]. The main goal is to decide the type of mobile inverted bottleneck convolution. To be more specific, MBConv layer is decided by the kernel size of the depthwise convolution $k$ and by the expansion ratio $e$. In particular, we consider layers with kernel sizes ${(3, 5)}$ and expansion ratios of ${(3, 6)}$. The backbone has 4 blocks each block consists of 4 groups of a point-wise 1*1 convolution, a k*k depthwise convolution, and a linear 1*1 convolution. To reduce the complexity of the problem, we only quantize weights and asume each layer within a group share the same quantization policy. The choices of quantization are limited within (4, 8, 16) bits.
\subsection{Training and Communication Efficiency}
For a fair comparison with FedAvg and FedNAS+EDD, we fisrt conduct simple experiments without hardware-aware preferences ($\alpha=1$ and $\beta=\gamma=0$) as FedAvg does not apply latency or model size analysis. As showed in Fig. 3 (left), FedAvg converges faster than FedNAS+EDD and FAQS due to no additional trainable architecture parameters. While both FedNAS+EDD and FAQS outperform FedAvg in final top 1 accuracy (88.3 88.2 and 85.1) with enough training rounds($ \sim$100 rounds). Meanwhile, FAQS demands less affordable search time than FedNAS+EDD due to the efficiency of weight-sharing super kernels and bit-sharing quantization. By plotting the average communication bits of FAQS curve on each round showed in Fig. 3 (middle), the total communication cost can be caculated on the area under the curves. As showed in Fig. 3 (right), FAQS achieves average reduction of 1.58x in communication bandwith per round compared with normal FedAvg and 4.51x in total communication compared with FedNAS+EDD framwork.

\subsection{Personalized Hardware-aware Preferences}
In order to demonstrate the ability of generating heterogeneous models for personalized hardware-aware preferences, we use 8 groups of  pareto loss with various $(\alpha,\beta,\gamma)$ for each node. By running through 150 rounds of FAQS, we record the accuracy curves and the scatterings on three clients' latency and model size. Client $A$ with (0.8,0.1,0.1) prefers accuracy much more than latency and model size, while client $B$ with (0.4,0.5,0.1) and $C$ with (0.4,0.1,0.5) are lat-prior and ms-prior respectively. As showed in Fig. 4 (left), the model searched in node $A$ outperforms $B$ and $C$ in accuracy. While both $B$ and $C$ yield high-performance models very near to corresponding Pareto boundries showed in Fig. 4 (right). The searched architectures and quantization policies are detailed in  Table \uppercase\expandafter{\romannumeral6} along with their performance and hardware-aware parameters. We give a brief analysis on the searched architectures and quantization policies. As client $A$ prefers accuracy much more than latency and model size, it tends to search for deep and wide architectures with long bit(16-b) quantizations. Latency is the main optimizing component for Client $B$. As a result, it prefers wide but shallow architectures as the GPU device is good at performing parallel computings. Preserving large kernels can make up for the accuracy loss introduced by shallow architectures. For the ms-prior client $C$, many layers adapt narrow and shallow block types to reduce model size. We have similar findings in [14] that both lat-prior and ms-prior models have quite shallow layers but long quantizaiton bit length in early stage, because feartures of early stages are relatively small and need higher resolution. Besides, both of them prefer larger MBConv and longer quantization bit in downsampled layers to prevent severe accuracy loss.

\section{conclusion}
In this paper, we have presented FAQS, the first joint neural architecture and quantization optimization framework in federate learning. FAQS reduces the communication cost by utilizing weight-sharing super kernels, bit-sharing quantization and masked transmission. Moreover, FAQS can yield heterogeneous hardware-aware models for various user preferences by setting different personlized pareto functions on accuracy, latency and model size. Our evaluation shows that FAQS can achieve average reduction of 1.58x in communication bandwith per round compared with normal FL framework and 4.51x compared with FL+NAS framwork.

\end{document}